\documentclass{article}
\usepackage{ijcai23}

\usepackage{times}
\usepackage{soul}
\usepackage{url}
\usepackage[hidelinks]{hyperref}
\usepackage[utf8]{inputenc}
\usepackage[small]{caption}
\usepackage{graphicx}
\usepackage{amsmath}
\usepackage{amsthm}
\usepackage{booktabs}
\usepackage{algorithm}
\usepackage{algorithmic}
\usepackage[switch]{lineno}

\title{Value-based Fast and Slow AI Nudging}

\author{
    Author Name
    \affiliations
    Affiliation
    \emails
    email@example.com
}

\author{
Marianna B. Ganapini$^1$ \and 
Francesco Fabiano$^2$ \and 
Lior Horesh$^3$\and 
Andrea Loreggia$^4$\and  
Nicholas Mattei$^5$\and  
Keerthiram Murugesan$^3$\and 
Vishal Pallagani$^6$\and
Francesca Rossi$^3$\and 
Biplav Srivastava$^6$\and 
Brent Venable$^7$
\affiliations
$^1$Union College \\
$^2$University of Parma\\
$^3$IBM Research\\
$^4$University of Brescia\\
$^5$Tulane University\\
$^6$Univ. of South Carolina\\
$^7$Institute for Human and Machine Cognition 
}

\begin{document}

\maketitle

\begin{abstract}
Nudging is a behavioral strategy aimed at influencing people's thoughts and actions. Nudging techniques can be found in many situations in our daily lives, and these nudging techniques can targeted at human fast and unconscious thinking, e.g., by using images to generate fear or the more careful and effortful slow thinking, e.g.,  by releasing information that makes us reflect on our choices. In this paper, we propose and discuss a value-based AI-human collaborative framework where AI systems nudge humans by proposing decision recommendations. Three different nudging modalities, based on when recommendations are presented to the human, are intended to stimulate human fast thinking, slow thinking, or meta-cognition. Values that are relevant to a specific decision scenario are used to decide when and how to use each of these nudging modalities. Examples of values are decision quality, speed, human upskilling and learning, human agency, and privacy. Several values can be present at the same time, and their priorities can vary over time. The framework treats values as parameters to be instantiated in a specific decision environment. 
\end{abstract}

\section{Introduction}

A nudge is a tool that aims to steer someone's course of action or thought in a certain direction. In their book "Nudge" \cite{nudgebook1}, Thaler and Sunstein focus on the behavioral aspect of human decision making and define a nudge as an element of a choice architecture, "understood as the background against which people make choices" \cite{nudgebook2}. This element "alters people’s behavior in a predictable way without forbidding any options”
(\cite{nudgebook2}: p 6). 
That is, nudges arguably preserve (a certain degree of) freedom of choice: they influence us without being fully manipulative or coercive (\cite{nudgepaper}: 489). Even if nudged, we can choose to opt-out fairly easily and take a different path. This is true even if we rarely fully recognize the influence of nudges on us and we find ourselves adopting a certain course of action without knowing that we have been influenced.  

In his book "Human Agency and Behavioral Economics", \cite{nudgebook2} refers to the fact that there are two macro-categories of nudges. Some nudges elicit certain emotions or habits in us and make us react accordingly. These nudges require little time and attention from us, they can be mindlessly adopted and followed. The fact that certain food is displayed more prominently in a cafeteria, for instance, makes it more likely that we will choose that food. This is because we have the disposition to opt for things that are more visible and convenient to reach so as to avoid wasting too much energy while choosing or acting. Hence, if we plan to encourage people to eat healthily, we need to put fruit and vegetables on the most easy-to-reach shelves, while fries and burgers are somewhere less visible and reachable.

As another example of mindless nudging, in certain countries, cigarette packages have some graphic warnings on them. These warnings are meant to scare people and discourage them from smoking \cite{nudgebook2}. This nudging strategy aims at triggering our emotions and thus engaging the fast and shallow thinking that leads us to adopt instinctive reactions. In psychology, this type of thinking has been called "thinking fast" or "System 1" thinking, from the work of Daniel Kahneman \cite{Kahneman}. It is a type of thinking that is mostly unconscious, automatic, based on fast and frugal heuristics, and often driven by emotions. As a result, this thinking may lead to biases, shortcuts, and reasoning errors.

When we think fast, we make decisions by reacting to environmental stimuli and generating actions out of past experiences in similar conditions. Thus, this type of thinking produces a limited range of possible outcomes and these outcomes can oftentimes be easily predicted. 
Therefore, nudging System 1 thinking means exploiting someone's fast but shallow thinking to motivate them to accept a particular course of action, e.g. not smoking, or choosing healthy food. Because it works mostly unconsciously, through an automatic and emotion-infused process, nudging directed at System 1 is seen as scarcely conducive to fully autonomous human choices: when subject to this influence we adopt a more passive stance and let the nudging guide us. 
 
\begin{figure}
    \centering
    \includegraphics[width=\linewidth]{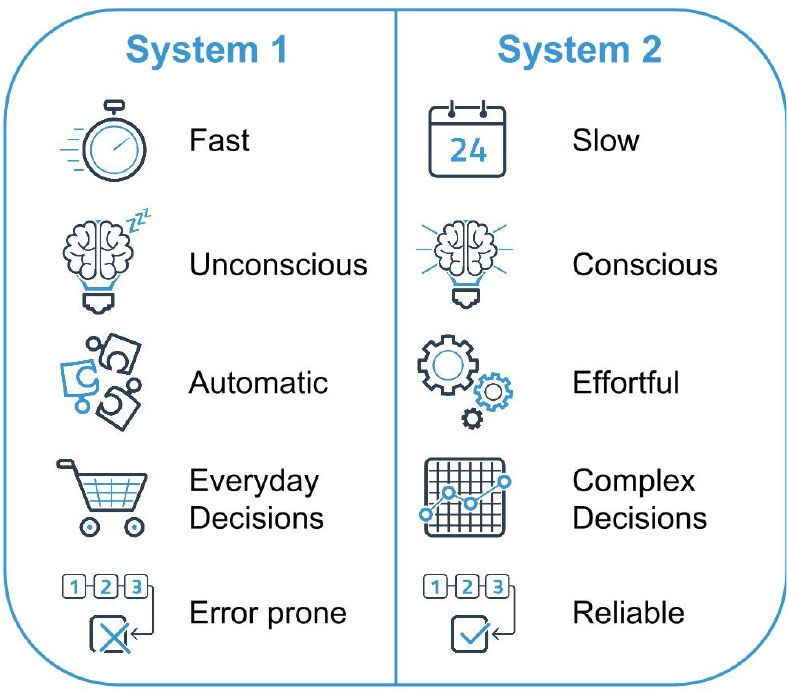}
    \caption{Thinking fast (System 1) and slow (System 2) in human decision making.}
    \label{fig:s1s2}
\end{figure}

A second type of nudging aims at engaging our "System 2" or "thinking slow" processes \cite{Kahneman}. This type of thinking is mostly conscious and under our control, and it engages rationality rather than relying on fast and frugal rules of thumb. When we think slow, we carefully reason about the problem to be solved, we devote all our attention to the reasoning process, and usually, we are more accurate in the result. Especially when conflicting solutions become available, or when the problem is recognized as too complex for System 1, System 2 kicks in and solves it with access to additional computational resources, full attention, and sophisticated logical reasoning. A typical example of a problem handled by System 2 is solving a complex arithmetic calculation, or a multi-criteria optimization problem. Figure \ref{fig:s1s2} shows the main features of both kinds of thinking.

There are nudges that can push us to adopt this more reflective stance and make choices that result from System 2. Sunstein calls it "educative" nudging because it encourages us to exercise our rationality. Others call this "cognitive" nudging \cite{edunudges}. 
Examples of this kind of nudging are the labels that we find on packaged food, which indicate the amount of calories, carbs, fats, sodium, and so on present in the food. These labels are nudges because they prompt us to be more reflective about the type of food we buy and to take a more informed and rational stance in relation to our dietary and nutritional goals. According to Sunstein \cite{nudgebook2}, this type of nudging motivates us to exercise our agency and develop autonomous decisions. Instead of encouraging us to indulge in our fast and shallow thinking, educational System 2 nudges push us to take charge of our choices. 

We experience nudging techniques extensively in our daily lives, for example by governments or other agencies, to push citizens towards behaviors that are more aligned with certain values that society has agreed upon and recognized as beneficial \cite{nudgebook1}. Nudges of all these kinds are typically accepted by most people \cite{nudgebook2}, since we feel that they help us improve our decisions or our beliefs towards values that we recognize as beneficial for us as individuals or as a society, such as health, well-being, privacy, transparency, and others.

Besides System 1 and System 2 nudges, in this paper, we are also considering a third type of nudging that pushes us towards meta-cognition and introspection, for example, to be reflective about our own level of confidence in solving a particular task. 
The concept of meta-cognition has been defined by \cite{Flavell} as the set of processes and mechanisms that allow a system to both monitor and control its own cognitive activities, processes, and structures, with the aim to improve the quality of the system’s decisions \cite{Cox}. Metacognition is able to assess our competence and knowledge in a domain, and the confidence we have in solving a particular task. We therefore model human metacognition as a higher-order representational process that requires cognitive effort and sophistication \cite{Carruthers}. 
Therefore, prompting metacognition is a way to make us introspect on possible gaps in our knowledge or lack of confidence in reaching a decision. Based on the results of its analysis, metacognition may decide whether to use System 1 or System 2 to solve a problem.
As an example of nudging metacognition, consider a question, that we often get them from our devices, about whether we want to set a reminder to complete a task for a specific deadline. This request pushes us to assess our own abilities: will we be able to remember what we have to do and for when? Will our memory likely go through a cognitive overload in the following days/weeks? Based on its assessment, metacognition may initiate the decision to set (or not) the reminder.

In our collaborative decision environment, that we call \emph{Fast and Slow Collaborative AI (FASCAI)}, the machine chooses how to interact with the human decision maker(s) to reach a final decision. Given a problem instance, the machine generates a decision
recommendation and presents it to the human decision maker, in a
way that stimulates the human’s appropriate modality of thinking, i.e., System 1, System 2, or metacognition. The conditions under which each of these three nudging mechanisms is chosen depend on appropriate values that are desired and relevant for the decision environment. 
In this paper we will consider values such as human upskilling, human agency, and decision quality. Other possible values to consider could be, for example, speed, human safety, and societal acceptance.

We are aware that the use of nudging techniques may also have harmful effects depending on why, when and towards whom those nudging techniques are used. Especially when machine-nudges are targeting protected categories (e.g. children), we should be extremely careful about the risks of adopting this technology \cite{AInudgerisk1,AInudgerisk2}.
We believe that the most innovative aspect of our nudging framework for AI-human collaboration stems exactly from recognizing and embracing the value-laden nature of this technology. While AI’s capabilities rapidly advance, it is fundamental to both include humans in the decision process and to ensure that certain values are protected and supported, to avoid undesired or uncontrolled outcomes \cite{RoMa19a,LoMaRoVe18a,rossi2019preferences}. This is why, in the framework proposed we introduce a flexible parametrization - based on relevant values - of our model. 

The FASCAI architecture provides a general nudge-based human-machine collaborative framework that can be flexibly instantiated by identifying appropriate values and priorities, tied to the decision environment’s parameters and that can evolve over time. 
A careful assessment of the parameters is especially important when decisions may be very high-stake and complex, and therefore require the combination of both humans' and AI’s capabilities to make sure that key goals and values are ensured and protected \cite{burton2019heart}. 
Without a transparent and agile approach to embedding values in human-AI systems, AI will not be considered trustworthy, especially in high-stake decision environments, and this will impact negatively its successful and beneficial adoption \cite{guidelinesHUMI}.

\textbf{Contributions:} The contributions of this paper can be summarized as follows: 
\begin{itemize}
    \item We introduce the notion of AI nudging in human-AI collaboration, leveraging knowledge and cognitive theories of how humans make decisions;
    \item We present the FASCAI architecture for human-AI collaboration via nudging, which is parametrized on relevant values to be supported in the collaborative decision making; 
    \item We present a specific instance of the FASCAI architecture, where the supported values are human agency, human upskilling, and decision quality;
    \item We define and discuss some of the research questions to be addressed both for the general FASCAI architecture and in the instance we present.
\end{itemize}

The novelty of this paper is the introduction of AI nudging that is informed by cognitive theories of human decision making, and values in the context of human-AI collaborative decision making, and in raising some research questions that we hope to work on with a wider multi-disciplinary community. We strongly believe that AI will be increasingly used to support human decisions in many areas, including human creativity and innovation. While this is very exciting and promising, we need to make sure that this collaboration between humans and machines protects and supports human and societal well-being and human-machine trust, besides ensuring system performance. This paper puts forward a proposal to achieve these ends. We hope to work with many other researchers, in a multidisciplinary environment, to discuss it, test it, expand it, generalize it, and adopt it.


\section{A Nudge-based Human-AI Collaborative Framework}


In this paper, we will draw from the insights on nudging, described in the previous section, to develop an AI technology that can help humans in making decisions, by leveraging the knowledge of how humans use fast and slow thinking modalities and metacognition. We believe that this approach offers advantages over other types of human-machine interaction \cite{HUMI,guidelinesHUMI,success1}.

In our collaborative decision environment, the machine chooses how to interact with the human decision maker(s) to reach a final decision, within the boundaries of values that humans (we) have defined. Given a problem instance, the machine generates a decision recommendation and presents it to the human decision maker, in a way that stimulates the human’s appropriate modality of thinking, with the aim of delivering a decision while supporting certain values. Examples of values in this context can be speed, human upskilling, human agency, and decision quality.

The specific recommendation to present to the decision maker can be influenced by the relative utilities of the various options.  In the simplest case, if the estimated utility of one decision is significantly better than all other alternatives, it becomes the selected recommendation.  If there are two (or more) alternatives that have comparable utility, the recommendation can be selected from the acceptable alternatives to either confirm the human choice or encourage thinking about an alternative choice.

We envision five human-machine decision modalities. 
Out of these five modalities, three of them adopt nudging mechanisms from the machine to the human, as the machine nudges the human to engage her System 1, her System 2, or her metacognition. In the other two modalities, either the machine or the human decides autonomously. 

\begin{figure}
    \centering
    \includegraphics[width=\linewidth]{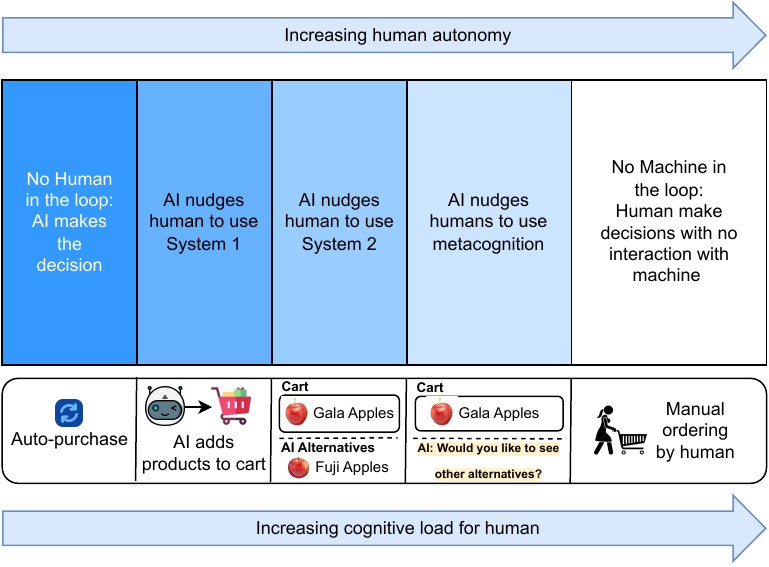}
    \caption{Five interaction modalities between AI and humans, leading to increased cognitive load and autonomy for humans from left to right. Also included is an example of the five interaction modalities, related to online product recommendation (bottom).}
    \label{fig:modalities}
\end{figure}

In Figure \ref{fig:modalities}, from left to right, we show interactions with an increasing cognitive load on the human, as well as their agency and autonomy. The more autonomous the human is, the more effort they will have to put into the decision. This is why we go from nudges to System 1 thinking, which is automatic and unconscious, to nudges for System 2 thinking, which requires effort and attention, to nudges for  metacognition, that 
requires not only thinking about the problem to be solved but also assessing one's own knowledge and capabilities. 

Figure \ref{fig:modalities} includes also an example of the five interaction modalities, focused on online product recommendation. When there is no human in the loop, the machine buys the product for us. In the System 1 nudging modality, the machine adds the product to our online cart, suggesting that we should buy it. With the System 2 nudging modality, we add a product to the cart and the machine proposes an alternative product; this forces us to think about which of the two we want to buy. With a metacognitive nudge, we add the product to the cart, and the machine asks us if we want to see alternative products. Finally, with a human-only decision environment, we just buy the product manually with no interaction with the machine.

\begin{figure}
    \centering
    \includegraphics[width=\linewidth]{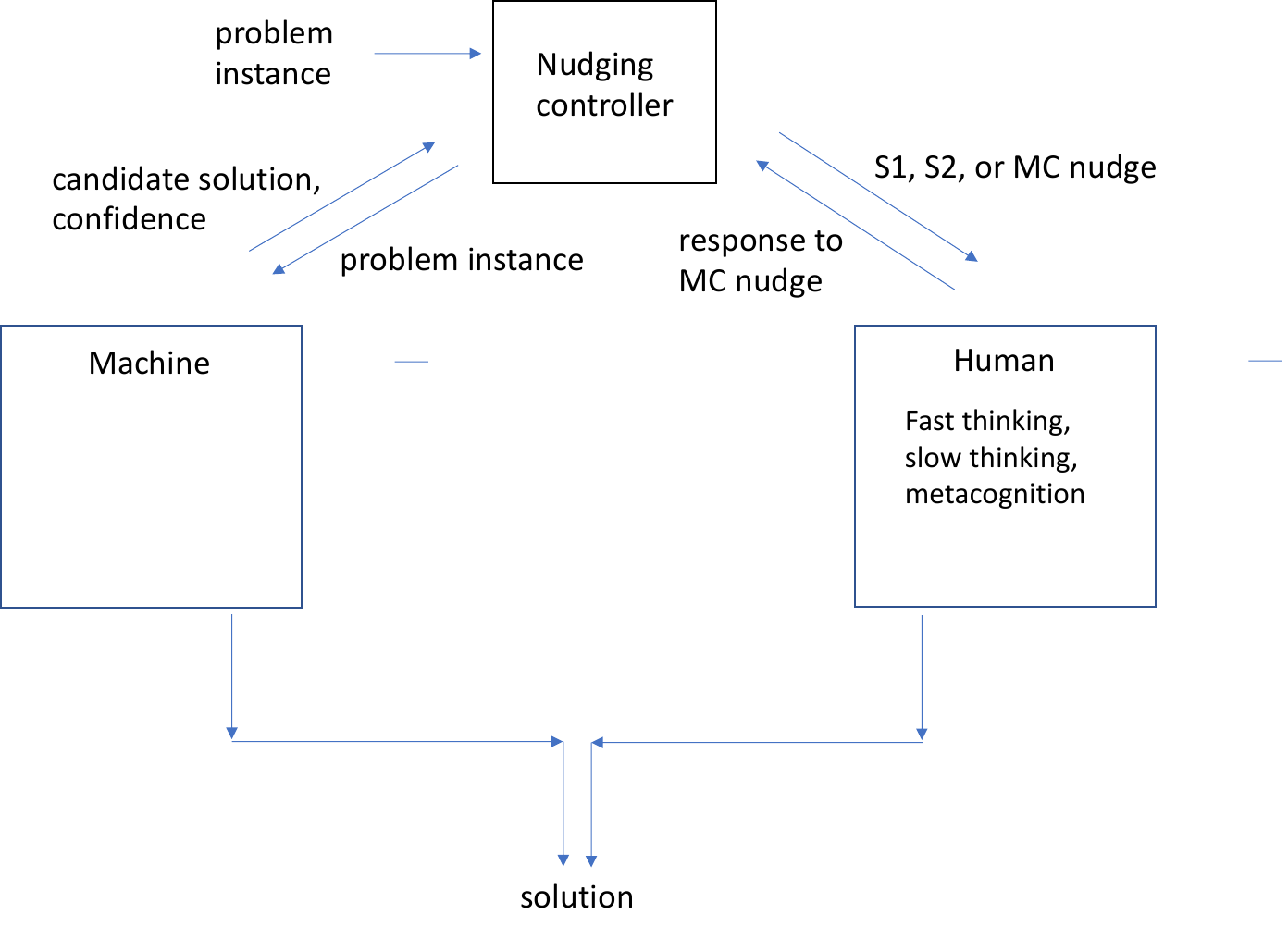}
    \caption{The FASCAI architecture, where a machine and a human interact via a nudging controller.}
    \label{fig:fascai}
\end{figure}

Besides the machine and the human, in our model, a third system receives the input problem instance and decides which interaction modality should be used, by activating the machine or the human in appropriate ways. We name this overall architecture FASCAI (for FAst and Slow Collaborative AI). Fig. \ref{fig:fascai} gives a schematic description of the system.

It is worth noting that, depending on the decision scenario, one may want to use only a subset of these interaction modalities. In particular, we envision that the autonomous decision by the AI system could be ruled out when the decision scenario is too risky, as for example is expected according to the European AI Act in the identified high-risk settings in the proposed legislation \cite{euaiact}, or when the risk of human safety is present.

\begin{figure}
    \centering
    \includegraphics[width=\linewidth]{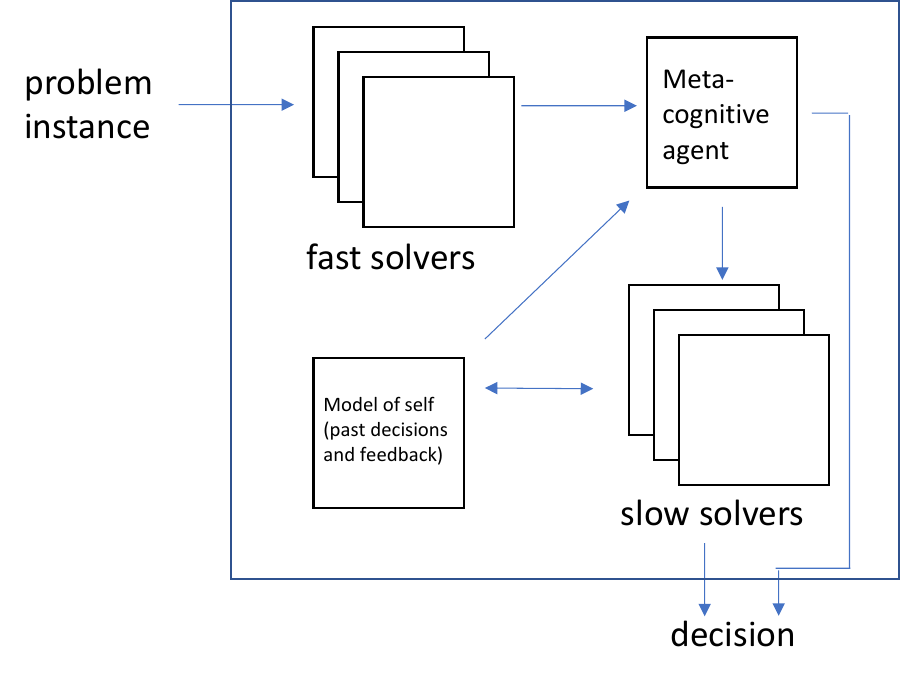}
    \caption{The SOFAI architecture, where a machine has both fast and slow solvers, plus a metacognitive agent to govern their use.}
    \label{fig:sofai}
\end{figure}

For this paper, we do not make any assumption on what technique the machine employs to generate the decision to recommend. However, it is worth noting that AI itself may work according to both fast data-driven approaches and slower logic-based reasoning methods, and can also learn how to navigate between the two in a way that minimizes time to decision and maximizes decision quality \cite{sofai,aaai2021-blue}. 
Figure \ref{fig:sofai} shows a schematic picture of the SOFAI architecture \cite{sofai}, that can be instantiated to build machines that have fast solvers, slow solvers, and a meta-cognitive agent that governs their use.

\section{How to Implement Nudges} 


Let us now consider the three nudging modalities in our framework: nudging for System 1, System 2, and metacognition, and discuss how they can be implemented.
The first nudging strategy is the one in which FASCAI decides to nudge the human to use her fast-thinking or System 1. To achieve this, the machine generates its decision recommendation and proposes it immediately to the human, without giving to the human any time to generate her own initial decision. 
This modality leverages the anchoring bias effect to make it difficult for the human to deviate from the machine recommendation, thus avoiding a deep System 2 thinking behavior. Because of both the automation and the anchoring bias effect, in this interaction modality, humans are most probably led to react to the machine by adopting its recommended decision, although they are free to choose to not do so \cite{Kahneman,time&anchoring,automationbias}.


As a second type of interaction, FASCAI may instead nudge the human towards adopting their slow thinking. To achieve this, the machine does not disclose the recommended decision right away, but rather gives it to the human only later, after having left enough time for the human to generate an initial decision. 
This will push the human to compare the two (possibly different) decisions and decide whether to revise her decision by using her slow thinking processes. This conjecture is based on studies that show that humans engage their System 2 thinking when faced with making a choice between contradictory options \cite{Kahneman,Pennycook}. 


The third type of machine nudging takes place when the machine encourages the human to use her own metacognition and thus to choose – in a deliberate, conscious way – which thinking modality to adopt or whether to ask for help \cite{Carruthers}. To achieve that, the machine waits for the human to generate an initial decision, and at that point, it gives the human the option to see the machine’s recommendation (while also disclosing its level of confidence and track record).
This question will likely trigger the human’s own metacognitive agent: the human is now nudged to check her own level of confidence and whether she could benefit from the machine’s help in making the final decision. This nudging modality gives humans great autonomy and encourages them to exercise their own agency in determining the best way to tackle the task at hand. 

\begin{figure}
    \centering
    \includegraphics[width=\linewidth]{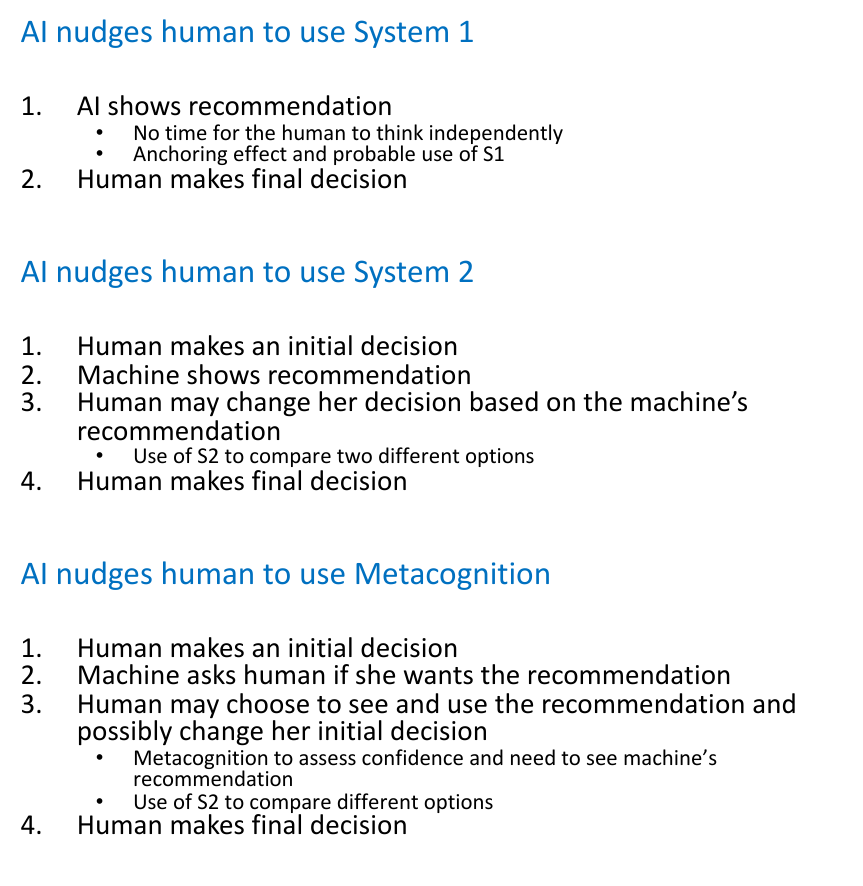}
    \caption{Nudges' implementation.}
    \label{fig:implementation}
\end{figure}

Figure \ref{fig:implementation} summarizes the proposed implementation of the three kinds of nudges.


\section{When to employ Each Nudging Modality: A Value-Based Approach}

Given the input problem instance, the collaboration controller of the FASCAI architecture decides the specific modality of interaction, based on the human's and the machine's past performance on similar problems, and the machine's own confidence in its recommendation. The choice of these two dimensions to structure the controller's decision depends on one fundamental value we aim to support with the human-machine interaction, which is {\em decision quality}: we want to deliver decisions that are superior to the decisions humans or machines would take by themselves. 

Let us assume, for sake of example, that we partition the machine confidence range in three intervals ({\em low}, {\em medium}, and {\em high} confidence). We, therefore, have six possible situations, based on three machine confidence levels and two comparison options for the human vs machine past performance. 

While we consider decision quality as a fundamental value to be supported in a decision making framework, there may be other values that are also very relevant for specific decision settings. These values should determine the allocation of the five interaction modalities onto the six confidence-performance combination scenarios.

For the decision scenarios we are describing, we assume that {\em human upskilling} and {\em human agency} are the other two values we want to support. That is, we would like humans to be in control of their decisions as much as possible, and to learn and improve their ability to think, reflect, and introspect while interacting with machines in making decisions. 
To achieve this, we allocate System 1 nudges only when really needed, preferring in all other cases System 2 and metacognition nudges.
More precisely, FASCAI will choose to adopt System 1 nudges when the machine is both highly confident in its suggestion and has a better track record on solving these types of tasks than humans. This indicates that the machine's suggestion is probably correct and that it is reasonable to nudge the human to adopt a specific behavior as we expect the human not to do well if left to decide fully autonomously. 
However, in the current framework, we limit human autonomy only in those situations in which the human performance has been significantly worse than the machine's, and we believe that the human could learn more by looking at what the machine does. Though this limits human's freedom of choice, we do not think that humans are completely deprived of their autonomy in those scenarios either. Even when nudged in their System 1, humans can always opt out and decide independently of what the machine recommends \cite{nudgepaper}. 

In contrast, System 2 nudging is used in two different scenarios. The first one takes place when the machine has a better performance than the human but for the current task its confidence in its own solution is low. In this situation, it is reasonable to rely on the human's slow thinking while also helping the human by showing them the machine's suggestion. We do not want here the machine to push the human to take a particular decision. In contrast, in this scenario, we offer the human the chance to improve on their past performance by encouraging them to consider the suggestion of the machine and confronting it with their own initial decision, so they can decide whether to follow the machine or not. In this situation, we may consider disclosing to the human that the machine's confidence is low to prevent automation bias to weigh on the human's final decision \cite{automationbias}.
Alternatively, the machine will nudge the human to adopt System 2 when the machine has a worse track record than the human but has high confidence in its current solution. Again, here we trigger the human's slow thinking by letting the human think about the problem first and reach an initial decision by themselves. Also, we expect that comparing their decision with the machine's suggestion will make the human more attentive when making the final decision. 

Since metacognition is especially demanding in terms of time and human cognitive load, we deem that only in certain conditions it is convenient to adopt this interaction modality, namely when the human is an expert and can reliably assess their own abilities \cite{decisionfatigue}. For this modality, the human has performed better than the machine in the past, and the machine has medium confidence in its recommended decision. In this situation, the machine offers its own solution after the human has reached an initial decision. Thus, we expect that humans will assess whether or not they may benefit from the machine's offer to help. 
This assumes that the human is capable of determining her confidence and therefore deciding if a machine can help (if the human's confidence is low), in which case the human will probably adopt System 2 to compare their candidate's decision with the machine's.


Our human-machine collaborative framework also includes two instances in which no nudging occurs. When the human has a superior track record and the machine is not confident in its own recommendation, the machine lets the human decide in full autonomy without nudging them in any way. Also, the machine will take the final decision without consulting the human in those cases when the machine is highly confident in its solution and the human's past performance is worse than the machine's. In this last scenario, the human will nevertheless be able to see what the machine does and the decisions it takes so that they can learn from the machine. 

\begin{figure}
    \centering
    \includegraphics[width=\linewidth]{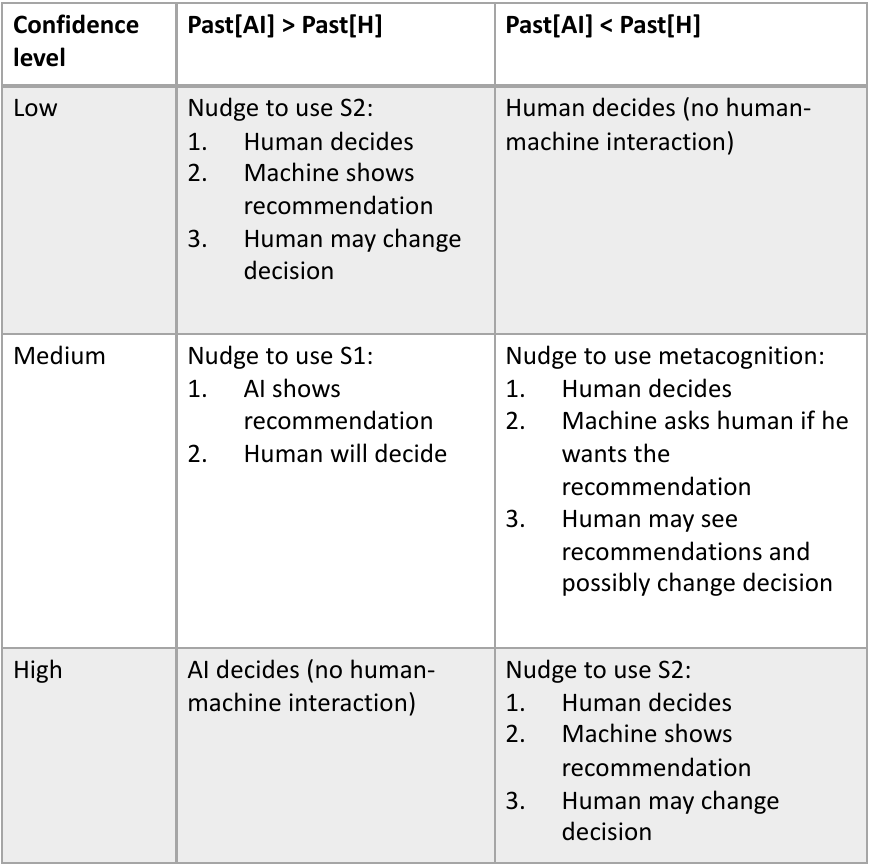}
    \caption{The nudges allocation table.}
    \label{fig:allocation}
\end{figure}

Figure \ref{fig:allocation} summarizes the allocation of the five interaction modalities to the various scenarios, depending on the machine's confidence and past performance of human vs machine.

We are building the FASCAI architecture in a way that 
can be parameterized by the values we think are relevant to the specific decision scenario. However, values can vary significantly depending on the decision environment. They include speed, human upskilling, decision quality, safety, resource consumption, human safety, and so on. Furthermore, for each situation often more than one value is relevant, so priorities should be defined and considered in filling the nudge allocation table shown above.
Values, and their priorities, can also vary within the same decision environment. For example, when time becomes a critical resource, speed may become more important than other values that were instead prioritized earlier. 
It is also reasonable to consider a feedback loop (from the final decisions to the nudging controller) that helps the nudging controller evolve over time: if the decision quality or the evidence for supporting other values decreases over time, we may want to modify the allocation table.
In conclusion, though in this example we have prioritized certain values over others, we believe that our overall model can be adapted to fit alternative values in different contexts. 

\section{Research Questions}

As mentioned in the introduction, this paper is intended to propose a general human-machine collaboration architecture and to explore its potential in embedding values in this domain, together with a wide research community. To this aim, we would also like to share some research questions about the FASCAI architecture, that we think will help us identify the best configurations for human-machine collaboration based on nudging.

Some of these questions are related to the assumptions we made in defining FASCAI, others refer to translating the nudging mechanism from a human to a machine environment, and the third category has to do with evaluating the quality of the values' embedding.


\subsection{FASCAI assumptions}

One of the main assumptions in our AI nudging approach is that just like in our everyday life also,  AI nudging creates an anchoring effect (that is, information coming from the machine leads humans to anchor their reasoning and decisions on that information). In fact, without a significant anchoring effect, we cannot expect machine nudging to achieve the effects we hope for in human reasoning and decisions. This is especially relevant for the System 1 nudges, where we assume that humans will probably follow the nudge.

\begin{quote}
    RQ1: Are AI nudges generating a significant anchoring effect?
\end{quote}

A related second question is about whether the machine anchoring effect is similar in strength (or less, or more) compared to the anchoring effect generated by human nudges. If so, we can safely adopt the extensive literature on human nudges for AI nudges.

\begin{quote}
    RQ2: Is the AI nudge anchoring effect comparable to the human one?
\end{quote}



\subsection{Human acceptance of AI nudges}

Another question is whether or not humans are inclined to accept machine-nudging at all, just like we accept human-generated nudges. It is possible that humans will tend to reject the suggestions coming from AI because they do not trust AI and are skeptical of its ability to deliver reliable results. Some initial experiments, on very specific decision scenarios, show significant effects on decision accuracy and a sort of prejudice against the machine  \cite{algoaversion1,algoaversion2,algoaversion3}. 

\begin{quote}
    RQ3: Are AI nudges accepted by humans as a human-machine collaboration mechanism? 
\end{quote}

A related question is whether humans accept more or less System 1 or System 2 AI nudges.

\begin{quote}
    RQ4: Are System 1 AI nudges accepted more or less than System 2 AI nudges? 
\end{quote}

\subsection{Value embedding}

In this paper, we have described both a human-AI collaborative framework (FASCAI) and one of its instances, that is expected to support decision quality, human upskilling, and human agency. 
For this instance, and in general for any other instance,
it is important to check that the desired values are indeed successfully embedded in the framework, that is, that they appear supported in the behavior of the system.

\begin{quote}
    RQ5: Are FASCAI's decisions better than humans' alone or AI's alone? 
\end{quote}

There is already some evidence that well-timed human-machine collaboration can lead to improved performance and decision-making \cite{success1,success2,success3}. However, those systems are not based on nudging. To study this for FASCAI, we plan to test the quality of the decisions adopted by FASCAI and compare them with the decisions of the human and the machine taken separately. 

\begin{quote}
    RQ6: In the FASCAI instance, are humans learning over time, more than if they are asked to make decisions by themselves?
\end{quote}

To show that adopting FASCAI supports human upskilling (that is, to answer question RQ6), we plan to conduct two sets of tests. First, we measure the human's performance at the relevant task before the interaction with the machine has taken place. After some runs of FASCAI, we then individually test the same human again: if they are performing better than they did initially, then this is an indication that they are acquiring new knowledge thanks to the collaboration with the machine.  

\begin{quote}
    RQ7: Is human agency preserved in FASCAI?
\end{quote}

To test this, we will study both System 1 and System 2 AI nudges to check the presence of human deviations from the machine nudge.

\section{Final Considerations and Future Work}

Nudging has been adopted and studied significantly in both human and human-machine settings \cite{humanfactor}. 
However, we are not aware of any other work that defines and fully adopts AI nudges that leverage the thinking fast and slow theory of human decision making. Moreover, we introduce here the possibility of adopting metacognitive nudges, a new perspective in the human-machine interaction \cite{HUMI}. Finally, and most importantly, we are not aware of any value-based framework for AI nudging (or collaborative AI). Therefore, the innovation of our work is in introducing and combining three complementary but related concepts within a general human-machine collaborative framework: values, AI nudging, and cognitive theories of human decision making.

We are now working to design appropriate testing frameworks for the research questions we have outlined in this paper, to implement them, and to assess the experimental results. We are also planning to combine the SOFAI architecture \cite{sofai} with the FASCAI framework described in this paper, to fully integrate the notion of thinking fast and slow in a human-machine collaborative environment.

\bibliographystyle{named}
\bibliography{biblio}

\begin{thebibliography}{}

\bibitem[\protect\citeauthoryear{Amershi \bgroup \em et al.\egroup
  }{2019}]{guidelinesHUMI}
Saleema Amershi, Dan Weld, Mihaela Vorvoreanu, Adam Fourney, Besmira Nushi,
  Penny Collisson, Jina Suh, Shamsi Iqbal, Paul~N Bennett, Kori Inkpen, et~al.
\newblock Guidelines for human-ai interaction.
\newblock In {\em Proceedings of the 2019 chi conference on human factors in
  computing systems}, pages 1--13, 2019.

\bibitem[\protect\citeauthoryear{Bahner \bgroup \em et al.\egroup
  }{2008}]{automationbias}
J~Elin Bahner, Anke-Dorothea H{\"u}per, and Dietrich Manzey.
\newblock Misuse of automated decision aids: Complacency, automation bias and
  the impact of training experience.
\newblock {\em International Journal of Human-Computer Studies},
  66(9):688--699, 2008.

\bibitem[\protect\citeauthoryear{Baudel \bgroup \em et al.\egroup
  }{2021}]{success1}
Thomas Baudel, Manon Verbockhaven, Victoire Cousergue, Guillaume Roy, and Rida
  Laarach.
\newblock Objectivaize: Measuring performance and biases in augmented business
  decision systems.
\newblock In {\em Human-Computer Interaction--INTERACT 2021: 18th IFIP TC 13
  International Conference, Bari, Italy, August 30--September 3, 2021,
  Proceedings, Part III 18}, pages 300--320. Springer, 2021.

\bibitem[\protect\citeauthoryear{Booch \bgroup \em et al.\egroup
  }{2021}]{aaai2021-blue}
Grady Booch, Francesco Fabiano, Lior Horesh, Kiran Kate, Jonathan Lenchner,
  Nick Linck, Andreas Loreggia, Keerthiram Murgesan, Nicholas Mattei, Francesca
  Rossi, and Biplav Srivastava.
\newblock Thinking fast and slow in {AI}.
\newblock In {\em Proceedings of the AAAI Conference on Artificial
  Intelligence}, volume~35, pages 15042--15046, 2021.

\bibitem[\protect\citeauthoryear{Burton \bgroup \em et al.\egroup
  }{2019}]{burton2019heart}
Emanuelle Burton, Kristel Clayville, Judy Goldsmith, and Nicholas Mattei.
\newblock The heart of the matter: patient autonomy as a model for the
  wellbeing of technology users.
\newblock In {\em Proceedings of the 2019 AAAI/ACM Conference on AI, Ethics,
  and Society}, pages 13--19, 2019.

\bibitem[\protect\citeauthoryear{Burton \bgroup \em et al.\egroup
  }{2020}]{algoaversion1}
Jason~W Burton, Mari-Klara Stein, and Tina~Blegind Jensen.
\newblock A systematic review of algorithm aversion in augmented decision
  making.
\newblock {\em Journal of Behavioral Decision Making}, 33(2):220--239, 2020.

\bibitem[\protect\citeauthoryear{Cabitza}{2019}]{algoaversion2}
Federico Cabitza.
\newblock Biases affecting human decision making in ai-supported second opinion
  settings.
\newblock In {\em Modeling Decisions for Artificial Intelligence: 16th
  International Conference, MDAI 2019, Milan, Italy, September 4--6, 2019,
  Proceedings 16}, pages 283--294. Springer, 2019.

\bibitem[\protect\citeauthoryear{Carruthers}{2014}]{Carruthers}
Peter Carruthers.
\newblock Two concepts of metacognition.
\newblock 2014.

\bibitem[\protect\citeauthoryear{Char \bgroup \em et al.\egroup
  }{2018}]{algoaversion3}
Danton~S Char, Nigam~H Shah, and David Magnus.
\newblock Implementing machine learning in health care—addressing ethical
  challenges.
\newblock {\em The New England journal of medicine}, 378(11):981, 2018.

\bibitem[\protect\citeauthoryear{Commission}{2021}]{euaiact}
European Commission.
\newblock European ai act, 2021.
\newblock
  \url{https://eur-lex.europa.eu/legal-content/EN/TXT/?uri=celex\%3A52021PC0206}.

\bibitem[\protect\citeauthoryear{Cox}{2005}]{Cox}
Michael~T Cox.
\newblock Metacognition in computation: A selected research review.
\newblock {\em Artificial intelligence}, 169(2):104--141, 2005.

\bibitem[\protect\citeauthoryear{Daniel}{2017}]{Kahneman}
Kahneman Daniel.
\newblock {\em Thinking, fast and slow}.
\newblock 2017.

\bibitem[\protect\citeauthoryear{Flavell}{1979}]{Flavell}
John~H Flavell.
\newblock Metacognition and cognitive monitoring: A new area of
  cognitive--developmental inquiry.
\newblock {\em American psychologist}, 34(10):906, 1979.

\bibitem[\protect\citeauthoryear{Ganapini \bgroup \em et al.\egroup
  }{2023}]{sofai}
M~Bergamaschi Ganapini, Murray Campbell, Francesco Fabiano, Lior Horesh, Jon
  Lenchner, Andrea Loreggia, Nicholas Mattei, Francesca Rossi, Biplav
  Srivastava, and Kristen~Brent Venable.
\newblock Thinking fast and slow in ai: the role of metacognition.
\newblock In {\em Machine Learning, Optimization, and Data Science: 8th
  International Workshop, LOD 2022, Certosa di Pontignano, Italy, September
  19--22, 2022, Revised Selected Papers, Part II}, pages 502--509. Springer,
  2023.

\bibitem[\protect\citeauthoryear{Hirshleifer \bgroup \em et al.\egroup
  }{2019}]{decisionfatigue}
David Hirshleifer, Yaron Levi, Ben Lourie, and Siew~Hong Teoh.
\newblock Decision fatigue and heuristic analyst forecasts.
\newblock {\em Journal of Financial Economics}, 133(1):83--98, 2019.

\bibitem[\protect\citeauthoryear{Khakurel and Blomqvist}{2022}]{success3}
Jayden Khakurel and Kirsimarja Blomqvist.
\newblock Artificial intelligence augmenting human teams. a systematic
  literature review on the opportunities and concerns.
\newblock In {\em Artificial Intelligence in HCI: 3rd International Conference,
  AI-HCI 2022, Held as Part of the 24th HCI International Conference, HCII
  2022, Virtual Event, June 26--July 1, 2022, Proceedings}, pages 51--68.
  Springer, 2022.

\bibitem[\protect\citeauthoryear{Lau \bgroup \em et al.\egroup
  }{2020}]{humanfactor}
Nathan Lau, Michael Hildebrandt, and Myounghoon Jeon.
\newblock Ergonomics in ai: designing and interacting with machine learning and
  ai, 2020.

\bibitem[\protect\citeauthoryear{Levy}{2017}]{edunudges}
Neil Levy.
\newblock Nudges in a post-truth world.
\newblock {\em Journal of medical ethics}, 43(8):495--500, 2017.

\bibitem[\protect\citeauthoryear{Loreggia \bgroup \em et al.\egroup
  }{2018}]{LoMaRoVe18a}
A.~Loreggia, N.~Mattei, F.~Rossi, and K.~B. Venable.
\newblock Value alignment via tractable preference distance.
\newblock In R.~V. Yampolskiy, editor, {\em Artificial Intelligence Safety and
  Security}, chapter~18. CRC Press, 2018.

\bibitem[\protect\citeauthoryear{Onnasch}{2015}]{success2}
Linda Onnasch.
\newblock Crossing the boundaries of automation—function allocation and
  reliability.
\newblock {\em International Journal of Human-Computer Studies}, 76:12--21,
  2015.

\bibitem[\protect\citeauthoryear{Parasuraman \bgroup \em et al.\egroup
  }{2000}]{HUMI}
Raja Parasuraman, Thomas~B Sheridan, and Christopher~D Wickens.
\newblock A model for types and levels of human interaction with automation.
\newblock {\em IEEE Transactions on systems, man, and cybernetics-Part A:
  Systems and Humans}, 30(3):286--297, 2000.

\bibitem[\protect\citeauthoryear{Pennycook \bgroup \em et al.\egroup
  }{2015}]{Pennycook}
Gordon Pennycook, Jonathan~A Fugelsang, and Derek~J Koehler.
\newblock What makes us think? a three-stage dual-process model of analytic
  engagement.
\newblock {\em Cognitive psychology}, 80:34--72, 2015.

\bibitem[\protect\citeauthoryear{Rastogi \bgroup \em et al.\egroup
  }{2020}]{time&anchoring}
Charvi Rastogi, Yunfeng Zhang, Dennis Wei, Kush~R Varshney, Amit Dhurandhar,
  and Richard Tomsett.
\newblock Deciding fast and slow: The role of cognitive biases in ai-assisted
  decision-making.
\newblock {\em arXiv preprint arXiv:2010.07938}, 2020.

\bibitem[\protect\citeauthoryear{Richard H.~Thaler}{2008}]{nudgebook1}
Cass R.~Sunstein Richard H.~Thaler.
\newblock Nudge: Improving decisions about health, wealth, and happiness, 2008.

\bibitem[\protect\citeauthoryear{Rossi and
  Loreggia}{2019}]{rossi2019preferences}
Francesca Rossi and Andrea Loreggia.
\newblock Preferences and ethical priorities: thinking fast and slow in {AI}.
\newblock In {\em Proceedings of the 18th international conference on
  autonomous agents and multiagent systems}, pages 3--4, 2019.

\bibitem[\protect\citeauthoryear{Rossi and Mattei}{2019}]{RoMa19a}
F.~Rossi and N.~Mattei.
\newblock Building ethically bounded {AI}.
\newblock In {\em Proceedings of the 33rd AAAI Conference on Artificial
  Intelligence (AAAI)}, 2019.

\bibitem[\protect\citeauthoryear{S{\ae}tra and Mills}{2022}]{AInudgerisk2}
Henrik~Skaug S{\ae}tra and Stuart Mills.
\newblock Psychological interference, liberty and technology.
\newblock {\em Technology in Society}, 69:101973, 2022.

\bibitem[\protect\citeauthoryear{Saghai}{2013}]{nudgepaper}
Yashar Saghai.
\newblock Salvaging the concept of nudge.
\newblock {\em Journal of medical ethics}, 39(8):487--493, 2013.

\bibitem[\protect\citeauthoryear{Smith and de
  Villiers-Botha}{2021}]{AInudgerisk1}
James Smith and Tanya de~Villiers-Botha.
\newblock Hey, google, leave those kids alone: Against hypernudging children in
  the age of big data.
\newblock {\em AI \& SOCIETY}, pages 1--11, 2021.

\bibitem[\protect\citeauthoryear{Sunstein}{2017}]{nudgebook2}
Cass~R Sunstein.
\newblock {\em Human agency and behavioral economics: Nudging fast and slow}.
\newblock Springer, 2017.

\end{thebibliography}

\end{document}